# Recognition and translation Arabic-French of Named Entities: case of the Sport places

**Abdelmajid Ben Hamadou**
MIRACL, University of Sfax
Sfax, Tunisia.

**Odile Piton**
Marin Mersenne, University Paris1
Panthéon Sorbonne, France.

**Héla Fehri**
LASELDI, University of Franche-Comte
Besançon, France.
MIRACL, University of Sfax
Sfax, Tunisia.

**Abstract**

The recognition of Arabic Named Entities (NE) is a problem in different domains of Natural Language Processing (NLP) like automatic translation. Indeed, NE translation allows the access to multilingual information. This translation doesn't always lead to expected result especially when NE contains a person name. For this reason and in order to ameliorate translation, we can transliterate some part of NE. In this context, we propose a method that integrates translation and transliteration together. We used the linguistic NooJ platform that is based on local grammars and transducers.

In this paper, we focus on sport domain. We will firstly suggest a refinement of the typological model presented at the MUC Conferences we will describe the integration of an Arabic transliteration module into translation system. Finally, we will detail our method and give the results of the evaluation.

## 1 Introduction

The recognition of named entities is an area of current research (Ehrmann 2008) given the proliferation of electronic documents exchanged through Internet and the need to process them by means of NLP tools. In this context, several works for identification and marking have been performed especially for Latin languages, and for English (Daille 2000). Language platforms dedicated to specific domains such as the medical field are developed (Hamon 2007). On the other hand, the translation of named entities from one language to another (Piton 2003), (Grass 2000) (Maurel 2007) opens new perspectives. It may be the basis of new applications such as, multilingual access to information (MAT), Document annotation/indexing and distance teaching of languages. Little work has been dedicated to Arabic NE (Mesfar 2007), (Mesfar 2008), (Fehri et al. 2008).

This article focuses on the recognition and translation from Arabic into French of NE. We are particularly interested in the names of athletic venues: stadiums, arenas, pools, tracks.... We propose a typological model specific to sport locations, refining the model presented at conferences MUC (MUC 6 and MUC 7) and an approach for recognition and translation from Arabic to French of the indicated NE. Based on rules, the implementation of this approach was performed using the NooJ platform [Silberstein 2005]. The rest of the article is organized as follows. We begin by detailing the typological model of NE of athletic venues. Then, we make a parallel between the grammars of NE in the two studied languages (Arabic and French). This parallel highlights the problems of Arabic-French translation for the context of study. Finally, we detail the approach proposed for the simultaneous extraction and translation of NE.

## 2   A typological model of NE of sport venues

Typing NE has been clearly specified for the first time at the conference MUC 6. The three basic types that have been defined under a hierarchy (Poibeau 2005) are: ENAMEX, NUMEX and TIMEX. ENAMEX includes three subcategories of proper names: People, Organizations and Locations; NUMEX includes of numerical expressions of percentages, quantities and monetary and TIMEX represents dates and durations.

Refinements have been proposed: Prolex project (Tran & Maurel 2006), whose purpose is to enable automatic processing of proper nouns, presented a modeling of ENAMEX type. This modeling has revealed the conceptual proper name and the prolexeme (i.e. a kind of super lemma of associated proper names) or family structured lexemes. Ontology has been specified taking into account the typology of Bauer (Grass 2002). This project has identified a two-level hierarchy with four selected hyper-types: Ethnonym (i.e. name of person, personality...), Toponym (i.e. place name : city, country ,..), Ergonym (i.e. article or manufactured product, ...) and Pragmonym (event and date,..). The names of athletic venues that we are dealing with are Toponyms.

The typological model that we propose is the result of a study of different forms of venues (stadiums, swimming pools, ice rink, ski slopes, ….) on corpus and lists of official names of sports venues available on the Internet for Arab countries (Tunisia, Algeria, Egypt, Saudi Arabia, Syria, ..) and Francophone countries (France, Belgium, Canada ,..). Our conclusion is that the typology proposed by the MUC conferences is unsuitable for our objective. We propose to refine it by adding new types related to the sub-category Locations (in MUC Hierarchy).

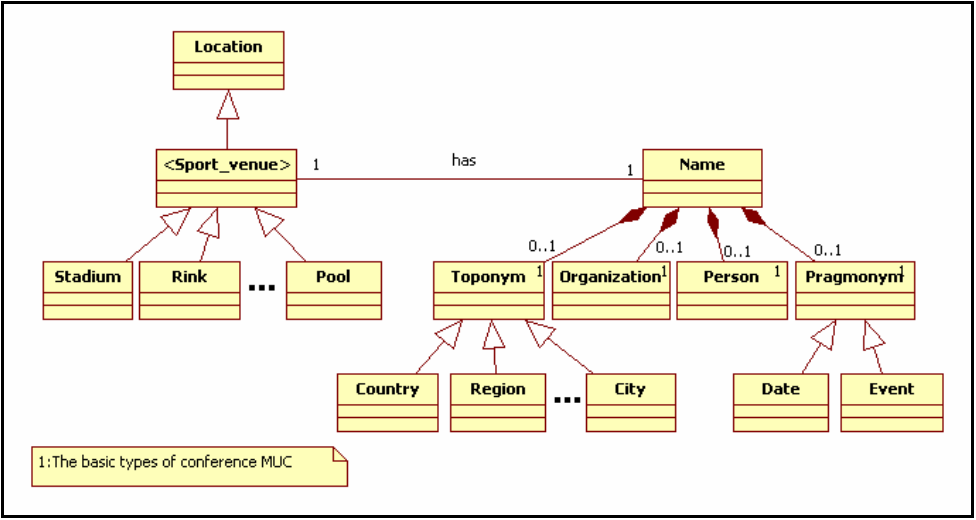

Figure1: The typological Model of NE sports venues

We particularly stress the compositionality and the recursivity of the NE of the sports venues. Indeed, a NE may include another one with which it is connected. This inclusion poses the problem of extracting the NE. Figure 3 gives two examples.

```
- <Examples>
  - <Example1>
      <SportVenue>استاد الملك فهد الدولي بالرياض = stade international roi Fahd de Ryadh</SportVenue>
    - <Categories>
        <SportVenueCategory>استاد = stade</SportVenueCategory>
        <Ethnonym>الملك فهد = roi Fahd</Ethnonym>
        <Adjective>الدولي = international</Adjective>
        <Toponym>الرياض = Ryadh</Toponym>
      </Categories>
    </Example1>
  - <Example2>
      <SportVenue>ملعب مدينة تشرين الرياضية = stade de la cité sportive Techrine</SportVenue>
    - <Categories>
        <SportVenueCategory>ملعب = stade</SportVenueCategory>
        <SportVenue>مدينة تشرين الرياضية = cité sportive Techrine</SportVenue>
      - <Categories>
          <SportVenueCategory>مدينة = cité</SportVenueCategory>
          <Pragmonym>تشرين = Techrine</Pragmonym>
          <Adjective>الرياضية = sportive</Adjective>
        </Categories>
      </Categories>
    </Example2>
</Examples>
```

Figure 2 Two instances of the typological model

## 3 Problems of recognition of Arabic NE

### 3.1 Agglutination problem

Arabic is an agglutinative language. Most words are formed by joining morphemes together. Indeed, textual forms are made up of the agglutination of prefixes (articles: definite article ال = the, prepositions: ل =for, conjunctions:= و and), and suffixes (linked pronouns) to the stems (inflected forms: لعبوا = لعـب+وا = they play). In general, to obtain the different decompositions of a textual form, a morphological grammar is needed.

### 3.2 Determination problem

Some constituents of sports venues names are always determined (in the case of adjectives). Others may be determined or not determined without rules governing these situations. This is the case of toponyms.

City name directly following the category:
ملعب صفاقس = Sfax stadium where the name is not determined and ملعب الرياض = Stadium Riyadh where the name is determined.

It is the same for countries and regions such as:
ملعب المغرب = stadium of Morocco and ملعب ليبيا = Libya stadium

This can also be met for anthroponyms:
ملعب شتيفي غراف = Steffi Graf Stadium: Undetermined
ملعب الشاذلي زويتن Stadium = the Chadli Zouiten (Sport Tunisian): determined

The treatment of this problem requires the inclusion of a feature in the dictionary indicating that the noun is or not determined.

### 3.3 The problem of proper names

Unlike the Latin languages, Arabic proper names do not begin by capital letters (upper case does not exist in Arabic). They are therefore difficult to identify. Also, their length is not known in advance, and may depend on the traditions of the region in which the person was born. Thus, in sports venues names, they can be written as a single name (ملعب الأسد = El-Asad stadium) or name and surname (ملعب الطيب المهيري= Taieb Mhiri stadium) or name and surname preceded by a title of nobility (ستاد الملك عبد الله = King Abdullah Stadium) or followed by "son of" (بن) of some regions (ستاد سحيم بن حمد = Sahim Bin Hamad Stadium). Furthermore, it is not possible to put in a dictionary all the names with all variants of writing. The transliteration could be a compromise solution because it is not easily to generalize.

### 3.4 The syntactic problems

The syntax of the NE in general, and those of sports venues in particular is rich and varied:
- The length of NE (or the number of components) cannot be known in advance and is variable. To complete the sense and not remove ambiguity, we tend to add an adjective (Municipal, Olympic, National,...) or the town name or the city name followed by the country name, etc.. الملعب الأولمبي بالمنزه = Stade Elmenzah ملعب مدينة الباسل الرياضية بدرعا = Stadium of the sport city El-Bacelli in Deraa.
- The same type of component can be found at different positions: it is mainly the case of the adjective that does not always follow the name to which it reports: ستاد عمان الدولي = Amman International Stadium الأستاد الوطني في بانكوك = National Stadium in Bangkok.
- The position of Toponyms is also variable. It can follow the category of sport venue or be at the end. استاد الملك فهد الدولي= King Fahd International Stadium ستاد حلب الدولي = Aleppo International Stadium.

## 4 Problems of Arabic-French translation of NE

The translation of NE from Arabic into French is not a trivial task. Several problems must be treated to produce a valid translation. We try to summarize these issues in the following points:
- Triggers Ambiguities of the source language. Example the word "stadium" which is a trigger name of athletic venues is also used to name sport club like الملعب التونسي = "Stade Tunisien"
- The feature of gender (male and female) is not always the same for the Arabic word and its equivalent in French. These features have an influence on the agreement within the NE. Example: The word مسبح is masculine, while its translation "piscine" is feminine.
- The translation of proper names and city names from one language to another depends on the existence of an exonym (like "Londres" pour London); otherwise, we should transliterate.
- Ambiguity of country names and capital in Arabic: e.g. the toponym تونس "Tunes" can be translated as "Tunisia" or "Tunis". It is also the case of الجزائر which can be translated as "Algiers or Algeria".
- The place of adjectives in the NE is not the same for both Languages: Example ملعب الملك عبد العزيز الأولمبي (the adjective is at the end) which translated into French by: King Abdelaziz Olympic Stadium (the adjective is in the middle).
- The use of determiners and prepositions in the order of words is not always the same. For the Arabic language, if the name of cities follows the category directly, there is no particle (ملعب موناكو), While in French we use 'of' better than 'in' (stadium of Monaco).
- The translation of dates is also problematic when it is expressed in the origin text as Hijri calendar (Hijri). Changes become necessary.

## 5 An approach to recognition and translation

Our approach of recognition and translation is rule oriented (an approach preferred to the statistical one when analysing texts). It is based on a balance between grammar and lexical resources. Te grammar indicates the composition rules of lexical components, in order to form a NE. The lexical resources correspond to one dictionary for each entity type indicated in the typological model pre-sented above. Each dictionary contains lemmas with the corresponding flexional model to generate all derived forms. The Arabic dictionaries entries contain, in addition, the corresponding French lemmas.

This list of the dictionaries for each language is as follows (see samples in Table1):
- Geographical names: City names and Country names
- Geographical categories (republic, jamahiriya, kingdom...)
- Adjectives associated with sports venues (e.g. olympic, national, municipal, sporting...).
- Personalities' names: Names of politicians, athletes...
- Categories of sports venues (stadium, swimming pool, ice rink...)
- Demonym (name/adjective for a resident of a locality or a country)
- Common name associated with sports such as place (fraternity, freedom, progress...)

To each entry of the indicated dictionary, a set of features is associated. The Arabic dictionaries have the features needed to identify a NE: function, Cat_Geo, Toponym, City, Country, Perso, Sport venue. The 'Function' feature concerns the function occupied by a person: e.g. Amir, president or king; the 'Cat_geo' feature indicates a geographical category: e.g. republic, city, region, the 'Toponyme' feature indicates a geographical place (a second feature is added to precise whether it is a city or a country). The 'Perso' feature indicates that it is a name of personality that can be a leader or an athlete.

The features indicated in the French dictionary entries are useful for translating the recognized elements of the Arabic NE: besides the classical features such as gender and number, we find the feature DETZ which characterizes the names without determiner, and the Apostrophe feature concerns the entries beginning with a vowel or a silent 'h', and requiring the use of the apostrophe after 'de' or after a determiner: either 'd' or 'l' (l'amitié). The French dictionary entries contain also the inflections of French lemmas. The use of flexion is necessary to know the appropriate translation of a given Arabic word.

| Arabic Dictionary | French Dictionary |
|---|---|
| **# Functions names** | **# names of sporting places** |
| مدير,N+Fonction+FLX=A1+FR=directeur | stade,N+FLX=Ballon |
| ملك,N+Fonction+FLX=ملك+FR=roi | amitié,N+apostrophe+FLX=ABH-O201 |
| **# names of geographical categories** | roi,N+FLX=ABH-Oroi |
| جمهورية,N+FLX=قارة+Cat_Geo+Toponyme+FR=république | Emir,N+apostrophe+FLX=ABH-Oemir |
| مملكة,N+FLX=قارة+Cat_Geo+Toponyme+FR=royaume | armée,N+apostrophe+FLX=Table |
| **# Geographical names** | **# adjectives** |
| تونس,N+PR+s+Pays+Toponyme+FR=Tunisie | municipal,A+FLX=76 |
| تونس,N+PR+s+Ville+Toponyme+FR=Tunis | olympique,A+FLX=31 |
| رياض,N+PR+s+Ville+Toponyme+FR=Riyadh | national,A+FLX=76 |
| **# Personalities' names** | **# names of geographical categories** |
| إليزابات,N+PR+Perso+f+s+FR=Elisabeth | république,N+FLX=Table |
| بورقيبة حبيب,N+PR+Perso+m+s+FR="Habib Bourguiba" | ville,N+FLX=Table |
| **# Name of sporting places (triggers)** | royaume,N+FLX=Table |
| مركز,N+LieuSport+FLX=ملعب+FR=centre | **# Personalities' names** |
| مسبح,N+LieuSport+FLX=ملعب+FR=piscine | Habib Bourguiba,N+PR+m+s |
| ملعب,N+LieuSport+FLX=ملعب+FR=stade | Elisabeth II,N+PR+f+s |
| **# Common nouns** | Fahd,N+PR+m+s |
| الأخوة,N+FLX=قارة+FR=amitié | **# Geographical Proper names** |
| جيش,N+FLX=جيش+FR=armée | Riyadh,N+PR+ DETZ+FLX=ABH-O201 |
| **# adjectives** | Sfax,N+PR+ DETZ+FLX=ABH-O201 |
| وطني,A+FLX=A1+FR=national | Tunis,N+PR+DETZ+FLX=ABH-O201 |
| أولمبي,A+FLX=A1+FR=olympique | Tunisie,N+PR+FLX=ABH-O201 |
| بلدي,A+FLX=A1+FR=municipal | Indonésie,N+apostrophe+PR+FLX=ABH-O201 |
| دولي,A+FLX=A1+FR=international | Jakarta,N+PR+ DETZ+FLX=ABH-O201 |
| **# demonym adjectives** | Maroc,N+PR+FLX=ABH-O199 |
| تونسي,A+Toponyme+FLX=A1+FR=tunisien | Bangkok,N+PR+ DETZ+FLX=ABH-O199 |
| مصري,A+Toponyme+FLX=A1+FR=égyptien | Jaka Baring,N+PR+ DETZ+FLX=ABH-O199 |

Table 1: Extracts of French and Arabic dictionaries

The local grammars for recognition and translation of sports venues use other more elementary grammars:
- Morphological Grammar for decomposition of Arabic agglutinated words (e.g. input : الأولمبي, output : ال + أولمبي )
- Inflectional grammars for both languages
- Grammar for recognizing dates that represent, in our context, historical events.

The process of recognition and translation of sports venues requires 3 basic steps:
1. Recognition of Arabic NE.
2. Translation (literally) of different components of the recognized NE into French and saving the generated associated features in a specific form.
3. Reorganisation of the generated components in accordance with the French grammar (possible change of word order, change / addition of specific connectors and agreement correction).

These steps can be implemented in two different ways: by executing them in cascade or by combining the three steps (1+2+3) into one. The advantage of the first alternative is that it is, linguistically, more portable than the second. But it requires the definition of an appropriate internal generic representation of the recognised NE. We chose the second alternative, particularly for reasons of implementation simplicity and our expertise of NooJ platform does not allow us to construct such representation and reuse it in the step 3.

The recognition and translation system that we propose is based on transducers implementing the grammars above indicated. The recognition process is directed by the existence of triggers which may be internal (i.e. part of the NE) or external (i.e. announcing a NE). The stop condition of the process is when it encountered a preposition (ب، في) followed by a city name or an unknown word by the graph.

The main transducer for the recognition and translation is given in Figure 3.

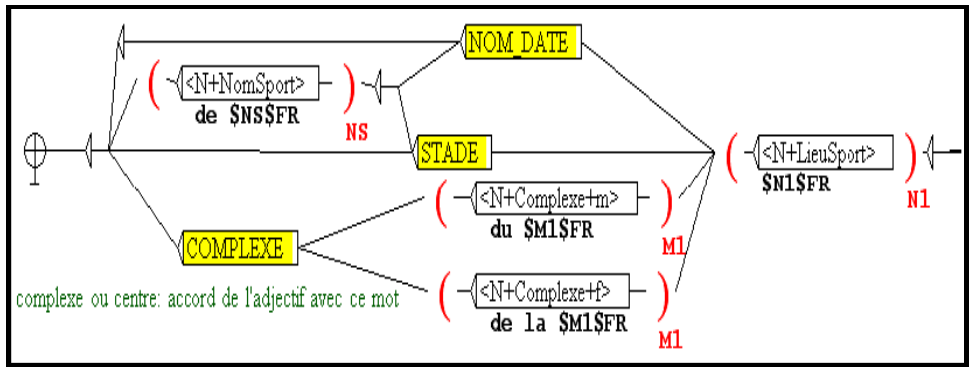

Figure 3: Main transducer for the recognition and translation of names of sports venues

The sub-graph NOM_DATE recognises a date. STADE recognise a NE beginning with a name of sporting places (triggers) and COMPLEXE recognises a complex NE.

As we stated above, adjectives pose a problem for both recognition and translation. To resolve this problem, we devoted a special transducer. This transducer allows the translation of the adjective before all the components that precede it in the source NE in order to respect the position of adjectives in the French noun phrase (see Figure 4).

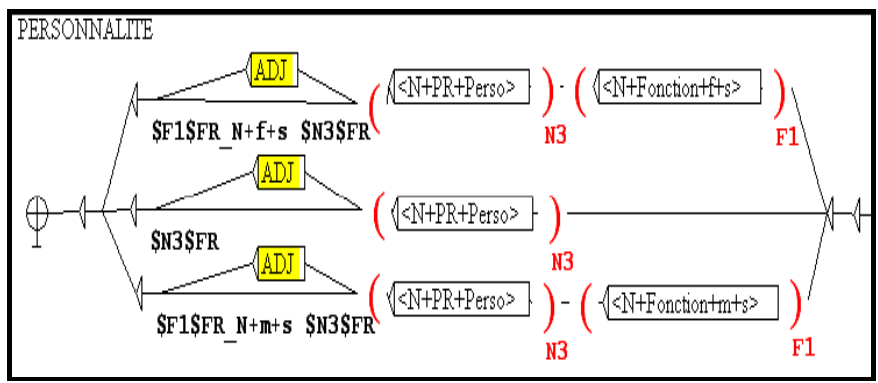

Figure 4: Transducer treating adjectives

The graph in Figure 5 shows an example of translating country names (Toponyms). This graph takes into account the presence / absence of determiner before the names of countries according to the gender (masc., fem.).

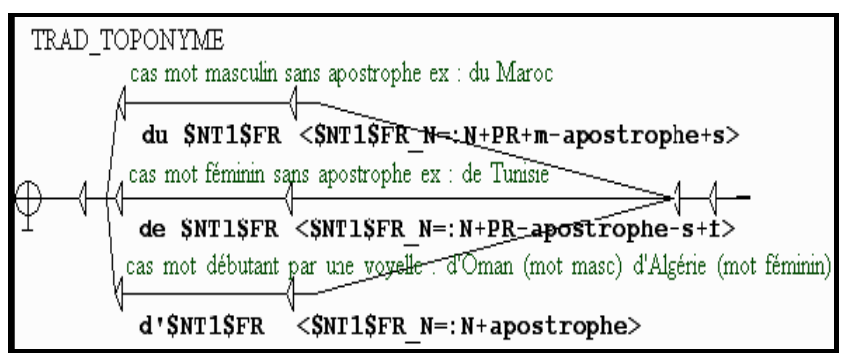

Figure 5: Transducer of translation of names of countries

## 6 Experimentation and Evaluation

To enumerate different forms of Sports venues names (patterns) and gather the used vocabulary, we collected a corpus (i.e. learning corpus) made up of journalistic articles and lists of official sports venues names available on the Internet. This corpus is made of a hundred texts of which 200 are sports venues NE.

The implemented system is composed of one main graph and 20 sub-graphs. These grammars use some dictionaries. Among used dictionaries, we can cite the following:

| Dictionaries | Number of entry |
| --- | --- |
| **Nouns of teams** | 5785 |
| **Nouns of sport** | 337 |
| **Nouns of country and capital** | 610 |

Let's note that we reused the Arabic dictionary of the first names (Mesfar, 2008). This dictionary contains 11852 entries.

For the system evaluation, we created a new corpus composed of the learning corpus and 200 pages from Wikipedia.

The figure 6 gives a sample execution of the proposed system given by Nooj platform on the indicated corpus. It represents the recognised Arabic sports venues NE with its corresponding French translation.

Figure 6: A sample of the concordance obtained for the evaluation corpus

The evaluation metrics we used are Recall, Precision and F_measure (2*P*R/(P+R)). Let's remember that the recall measures the quantity of relevant responses of the system compared to the ideal number of responses; Precision is the number of relevant responses of the system among all the responses he gave and the F-measure is a combination of Precision and Recall for penalizing the very large inequalities between these two measures. The values obtained in the evaluation of our work are:

|  | *Precision* | *Recall* | *F-measure* |
| --- | --- | --- | --- |
| *Sports venues* | 69% | 67% | 68% |

The obtained result value can be explained by the lack of standards for writing proper names, especially those Transliterated from foreign proper names and the omission of triggers in some sports venues named entity. For example, in some cases, we find some sports venues preceded by a trigger word (e.g. ملعب الطيب المهيري Taieb Mhiri stadium). Subsequently, in an other appearance of the same sport venue name, the trigger word is omitted (e.g. الطيب المهيري Taieb Mhiri), for the reason to avoid repetition and long sentences.

## 7    Conclusion

In this article we have presented an approach for recognition and translation from Arabic into French of sports venues names. We have particularly noted the recognition problems of which some are spe-cific to the Arabic language. These problems have been largely resolved, but some merit special consideration. In particular, we stress on the transliteration of the proper names and the abbreviations and acronyms. Transliteration allows reducing the dictionary size of proper names. One solution has been proposed for this problem, but the results are not yet satisfactory because of the diversity of the transliteration standards and systems (Fehri et al. 2009).

About methodology, we are studying the separation of the recognition step from the translation step and defining an internal representation of the analysed NE. The objective is to reuse the recognition step for the translation of the Arabic EN into languages other than French.